\documentclass[tablecaption=bottom,pmlr]{jmlr}

\jmlrvolume{123}
\jmlryear{2020}
\jmlrworkshop{NeurIPS 2019 Competition and Demonstration Track}

\title[Large regression coefficients may predict causal links better than small p-values]{Causal structure learning from time series: Large regression coefficients may predict causal links better in practice than small p-values}
\author{%
\Name{Sebastian Weichwald} \Email{sweichwald@math.ku.dk}\\
\Name{Martin E Jakobsen} \Email{m.jakobsen@math.ku.dk}\\
\Name{Phillip B Mogensen} \Email{pbm@math.ku.dk}\\
\Name{Lasse Petersen} \Email{lp@math.ku.dk}\\
\Name{Nikolaj Thams} \Email{thams@math.ku.dk}\\
\Name{Gherardo Varando} \Email{gherardo.varando@math.ku.dk}\\
\addr Copenhagen Causality Lab, Department of Mathematical Sciences, University of Copenhagen}

\editors{Hugo Jair Escalante and Raia Hadsell.}

\usepackage{pgf}
\usepackage{xfrac}
\usepackage{algorithm2e}
\usepackage{float}

\definecolor{tidy}{RGB}{11,102,35}
\definecolor{factor}{RGB}{161,40,48}

\usepackage{bm}

\SetCommentSty{mycommfont}

\begin{document}

\setcounter{page}{27}

\maketitle

\begin{abstract}
In this article, we describe the algorithms for causal structure learning from time series data that won the Causality 4 Climate competition at the Conference on Neural Information Processing Systems 2019 (NeurIPS).
We examine how our combination of established ideas achieves competitive performance on semi-realistic and realistic time series data exhibiting common challenges in real-world Earth sciences data.
In particular, we discuss
a) a rationale for leveraging linear methods to identify causal links in non-linear systems,
b) a simulation-backed explanation as to why large regression coefficients may predict causal links better in practice than small p-values and
thus why normalising the data
may sometimes hinder causal structure learning.

For benchmark usage, we detail the algorithms here and provide implementations at \\\href{https://github.com/sweichwald/tidybench}{github.com/sweichwald/\textcolor{tidy}{tidybench}}.
We propose the presented competition-proven methods for baseline benchmark comparisons to guide the development of novel algorithms for structure learning from time series.
\end{abstract}

\section{Introduction}\label{sec:introduction}

Inferring causal relationships from large-scale observational studies is an
essential aspect of modern climate science~\citep{runge2019inferring, runge2019detecting}.
However, randomised studies and controlled interventions
cannot be carried out, due to both ethical and practical reasons.
Instead, simulation studies based on climate models are state-of-the-art to study
the complex patterns present in Earth climate systems~\citep{ipcc2013}.

Causal inference methodology can integrate and
validate current climate models and can be used to probe cause-effect relationships between
observed variables.
The Causality 4 Climate (C4C) NeurIPS competition~\citep{runge2020neurips} aimed to further the understanding and development of methods
for structure learning from time series data exhibiting common challenges in and properties of realistic weather and climate data.

\paragraph{Structure of this work}
Section~\ref{sec:structurelearning} introduces the structure learning task considered.
In Section~\ref{sec:winningalgorithms}, we describe our winning algorithms.
With a combination of established ideas, our algorithms achieved competitive performance on semi-realistic data across all $34$ challenges in the C4C competition track.
Furthermore, at the time of writing, our algorithms lead the rankings for all hybrid and realistic data set categories available on the \href{http://causeme.net}{CauseMe.net} benchmark platform which also offers additional synthetic data categories~\citep{runge2019inferring}.
These algorithms---which can be implemented in a few lines of code---are built on simple methods, are computationally efficient, and exhibit solid performance across a variety of different data sets.
We therefore encourage the use of these algorithms as baseline benchmarks and guidance of future algorithmic and methodological developments for structure learning from time series.

Beyond the description of our algorithms, we aim at providing intuition that can explain the phenomena we have observed throughout solving the competition task.
First, if we \emph{only} ask whether a causal link exists in some non-linear time series system, then we may sidestep the extra complexity of explicit non-linear model extensions (cf.\ Section~\ref{sec:nonlinearbylinear}).
Second, when data has a
meaningful natural scale, it may---somewhat unexpectedly---be advisable to forego data normalisation and to use raw (vector auto)-regression coefficients instead of p-values to assess whether a causal link exists or not (cf.\ Section~\ref{sec:coeffvspval}).

\section{Causal structure learning from time-discrete observations}\label{sec:structurelearning}
The task of inferring the causal structure from observational data is often referred to as `causal discovery' and was pioneered by \citet{Pearl2009} and \cite{Spirtes2000}.
Much of the causal inference literature is concerned with structure learning from
independent and identically distributed (iid) observations.
Here, we briefly review some aspects and common assumptions for causally modelling time-evolving systems.
More detailed and comprehensive information can be found in the provided references.

\paragraph{Time-discrete observations}
We may view the discrete-time observations as arising from an underlying continuous-time causal system~\citep{Peters2020}.
While difficult to conceptualise, the correspondence between structural causal models and differential equation models can be made formally precise~\citep{mooij2013from,rubenstein2016from,bongers2018random}.
Taken together, this yields some justification for modelling dynamical systems by discrete-time causal models.

\paragraph{Summary graph as inferential target}
It is common to assume a time-homogeneous causal structure such that the dynamics of the observation vector $X$ are governed by $X^{t} := F(X^{{\operatorname{past}(t)}},N^t)$
where the function $F$
determines the next observation based on past values $X^{\operatorname{past}(t)}$ and the noise innovation $N^t$.
Here, structure learning amounts to identifying the summary graph with adjacency matrix $A$ that summarises the causal structure in the following sense:
the $(i,j)^\text{th}$ entry of the matrix $A$ is $1$ if $X_i^{\operatorname{past}(t)}$ enters the structural equation of $X_i^t$ via the $i^{th}$ component of $F$ and $0$ otherwise.
If $A_{ij}=1$, we say that ``$X_i$ causes $X_j$''.
While summary graphs can capture the existence and non-existence of cause-effect relationships, they do in general not correspond to a time-agnostic structural causal model that admits a causal semantics consistent with the underlying time-resolved structural causal model~\citep{rubenstein2017causal,janzing2018structural}.

\paragraph{Time structure may be helpful for discovery}

In contrast to the iid setting, the Markov equivalence class of the summary graph induced by the structural equations of a dynamical system is a singleton when assuming causal sufficiency and no instantaneous effects~\citep{peters2017elements,wengel2019markov}.
This essentially yields a justification and a constraint-based causal inference perspective on Wiener-Granger-causality \citep{wiener1956theory,granger1969investigating,peters2017elements}.

\paragraph{Challenges for causal structure learning from time series data}

Structure learning from time series is a challenging task hurdled by further problems such as time-aggregation, time-delays, and time-subsampling.
All these challenges were considered in the C4C competition and are topics of active research~\citep{danks2013learning,hyttinen2016causal}.

\section{The \textcolor{tidy}{ti}me series \textcolor{tidy}{d}iscover\textcolor{tidy}{y} \textcolor{tidy}{bench}mark \texttt{(\textcolor{tidy}{tidybench})}: Winning algorithms}
\label{sec:winningalgorithms}

We developed four simple algorithms,
\begin{altdescription}{SLARAC}
\setlength\itemsep{0em}

\item[\texttt{SLARAC}] Subsampled Linear Auto-Regression Absolute Coefficients (cf.\ Alg.~\ref{alg:varvar})

\item[\texttt{QRBS}] Quantiles of Ridge regressed Bootstrap Samples~(cf.\ Alg.~\ref{alg:QRBS})

\item[\texttt{LASAR}] LASso Auto-Regression

\item[\texttt{SELVAR}] Selective auto-regressive model

\end{altdescription}
which came in first in 18 and close second in 13 out of the 34 C4C competition categories and won the overall competition~\citep{runge2020neurips}.
Here, we provide detailed descriptions of the \texttt{SLARAC} and \texttt{QRBS} algorithms.
Analogous descriptions for the latter two algorithms and implementations of all four algorithms are available at \href{https://github.com/sweichwald/tidybench}{github.com/sweichwald/\textcolor{tidy}{tidybench}}.

All of our algorithms output an edge score matrix that contains
for each variable pair $(X_i,X_j)$ a score that reflects how likely it is that the edge $X_i \to X_j$ exists.
Higher scores correspond to edges that are inferred to be more likely to exist than edges with lower scores, based on the observed data.
That is, we rank edges relative to one another but do not perform hypothesis tests for the existence of individual edges.
A binary decision can be obtained by choosing a cut-off value for the obtained edge scores.
In the C4C competition, submissions were compared to the ground-truth cause-effect adjacency matrix and assessed based on the achieved ROC-AUC when predicting which causal links exist.

The idea behind our algorithms is the following:
regress present on past values and inspect the regression coefficients to decide whether one variable is a Granger-cause of another.
\texttt{SLARAC} fits a VAR model on bootstrap samples of the data each time choosing a random number of lags to include;
\texttt{QRBS} considers bootstrap samples of the data and Ridge-regresses time-deltas $X(t) - X(t-1)$ on the preceding values $X(t-1)$;
\texttt{LASAR} considers bootstrap samples of the data and iteratively---up to a maximum lag---LASSO-regresses the residuals of the preceding step onto values one step further in the past and keeps track of the variable selection at each lag to fit an OLS regression in the end with only the selected variables at selected lags included;
and \texttt{SELVAR} selects edges employing a hill-climbing procedure based on the
leave-one-out residual sum of squares and finally scores the selected
edges with the absolute values of the regression coefficients.
In the absence of instantaneous effects and hidden confounders, Granger-causes are equivalent to a variable's causal parents~\cite[Theorem 10.3]{peters2017elements}.
In Section~\ref{sec:coeffvspval}, we argue that the size of the regression coefficients may in certain scenarios be more informative about the existence of a causal link than standard test statistics for the hypothesis of a coefficient being zero.
It is argued that for additive noise models, information about the causal ordering may be contained in the raw marginal variances. In test statistics such as the F- and T-statistics, this information is lost when normalising by the marginal variances.

\begin{figure}
\begin{algorithm2e}[H]
\SetAlgoLined
\SetKwInOut{input}{Input\,}
\SetKwInOut{params}{Parameters\,}
\SetKwInOut{output}{Output\,}
\input{%
Data $\boldsymbol{X}$ with $T$ time samples $\boldsymbol{X}(1),\dots,\boldsymbol{X}(T)$ over $d$ variables.
}
\params{%
Max number of lags, $L \in \mathbb{N}$.\\
Number of bootstrap samples, $B \in \mathbb{N}$.\\
Individual bootstrap sample sizes, $\{v_1, \dots, v_B\}$.
}
\output{A $d\times d$ real-valued score matrix, $\widehat{A}$.}
\vspace{11pt}
\textbf{Initialise} $A_{\operatorname{full}}$ as a $d\times d L$ matrix of zeros and $\widehat{A}$ as an empty $d\times d$ matrix\;
\For{$b=1,\dots, B$}{%
$\text{lags} \leftarrow $ random integer in $\{1,\dots, L\}$\;
Draw a bootstrap sample $\{t_{1},\dots,t_{v_b}\}$ from $\{\operatorname{lags}+1, \dots, T\}$ with replacement\;
$\boldsymbol{Y}^{(b)}\leftarrow (\boldsymbol{X}(t_{1}),\dots \boldsymbol{X}(t_{v_b}))$\;
$\boldsymbol{X}_{\text{past}}^{(b)} \leftarrow \begin{pmatrix}
    \boldsymbol{X}(t_{1} - 1) & \cdots & \boldsymbol{X}(t_{1} - \text{lags}) \\
    \vdots & \ddots & \vdots \\
    \boldsymbol{X}(t_{v_b} - 1) & \cdots & \boldsymbol{X}(t_{v_b} - \text{lags})
\end{pmatrix}$\;
Fit OLS estimate $\beta$ of regressing $\boldsymbol{Y}^{(b)}$ onto $\boldsymbol{X}_{\text{past}}^{(b)}$\;
Zero-pad $\beta$ such that $\dim \beta = d\times dL$\;
$A_{\operatorname{full}} \leftarrow A_{\operatorname{full}} + |\beta|$\;
}
Aggregate $(\widehat{A})_{i,j} \leftarrow \operatorname{max}((A_{\operatorname{full}})_{i,j + 0\cdot d}, \dots ,(A_{\operatorname{full}})_{i,j+L\cdot d})$ for every $i,j$\;
\textbf{Return:} Score matrix $\widehat{A}$.
\caption{Subsampled Linear Auto-Regression Absolute Coefficients (\texttt{SLARAC})}
\label{alg:varvar}
\end{algorithm2e}
\vspace*{2em}
\begin{algorithm2e}[H]
\caption{Quantiles of Ridge regressed Bootstrap Samples (\texttt{QRBS})}
\label{alg:QRBS}
\SetKwInOut{input}{Input\,}
\SetKwInOut{params}{Parameters\,}
\SetKwInOut{output}{Output\,}
\input{%
Data $\boldsymbol{X}$ with $T$ time samples $\boldsymbol{X}(1),\dots,\boldsymbol{X}(T)$ over $d$ variables.
}
\params{%
Number of bootstrap samples, $B \in \mathbb{N}$.\\
Size of bootstrap samples, $v \in \mathbb{N}$.\\
Ridge regression penalty, $\kappa \geq 0$.\\
Quantile for aggregating scores, $q \in [0, 1]$.
}
\output{A $d\times d$ real-valued score matrix, $\widehat{A}$.}
\vspace{11pt}
\For{$b=1, \dots, B$}{%
Draw a bootstrap sample $\{t_{1}, \dots, t_{v}\}$ from $\{2,\dots, T\}$ with replacement\;
$\boldsymbol{Y}^{(b)} \leftarrow (\boldsymbol{X}(t_{1})-\boldsymbol{X}(t_{1}-1),\dots, \boldsymbol{X}(t_{v})-\boldsymbol{X}(t_{v}-1))$\;
$\boldsymbol{X}^{(b)} \leftarrow (\boldsymbol{X}(t_{1}-1),\dots, \boldsymbol{X}(t_{v}-1))$\;
Fit a ridge regression of $\boldsymbol{Y}^{(b)}$ onto $\boldsymbol{X}^{(b)}$:
$\widehat{A}_b = \arg\min_{A} \|\bm{Y}^{(b)} - A\bm{X}^{(b)}\| + \kappa \|A\|$\;
}
Aggregate $\widehat{A} \leftarrow q^{th} \text{ element-wise quantile of } \{|\widehat{A}_1|, \dots, |\widehat{A}_B|\}$\;
\textbf{Return} Score matrix $\widehat{A}$.
\end{algorithm2e}
\end{figure}

\section{Capturing non-linear cause-effect links by linear methods}\label{sec:nonlinearbylinear}

We explain the rationale behind our graph reconstruction algorithms and how they may capture non-linear dynamics despite being based on linearly regressing present on past values.
For simplicity we will outline the idea in a multivariate regression setting with additive noise, but it
extends to the time series setting by assuming time homogeneity.

Let $N, X(t_1), X(t_2) \in \mathbb{R}^d$ be random variables such that $X(t_2) := F(X(t_1)) + N$ for some differentiable function
$F = (F_1, \dots, F_d) : \mathbb{R}^d \to \mathbb{R}^d$. Assume that $N$ has mean zero, that
it is independent from $X(t_1)$, and that it has mutually independent components.
For each $i, j = 1, \dots, d$ we define the quantity of interest
  \begin{align*}
  \theta_{ij} = \mathbb{E} \left|\partial_i F_j\left(X(t_1)\right)\right|,
  \end{align*}
such that $\theta_{ij}$ measures the expected effect from $X_i(t_1)$ to $X_j(t_2)$.
We take the matrix $\Theta = \left(\bm{1}_{\theta_{ij}>0}\right)$ as the adjacency matrix of the summary graph
between $X(t_1)$ and $X(t_2)$.

In order to detect regions with non-zero gradients of $F$ we create bootstrap samples $\mathcal{D}_1, \dots, \mathcal{D}_B$.
On each bootstrap sample $\mathcal{D}_b$ we obtain the regression coefficients $\widehat{A}_b$ as estimate of the directional derivatives by a (possibly penalised) linear
regression technique.
Intuitively, if $\theta_{ij}$ were zero, then on any bootstrap sample we would obtain a small non-zero contribution.
Conversely, if $\theta_{ij}$ were non-zero, then we may for some bootstrap samples obtain a linear fit of $X_j(t_2)$ with large absolute regression coefficient for $X_i(t_1)$.
The values obtained on each bootstrap sample are then aggregated by, for example, taking the average of the absolute regression coefficients $\widehat{\theta}_{ij} = \frac{1}{B} \sum_{b=1}^B \left|(\widehat{A}_b)_{ij}\right|$.

This amounts to searching the predictor space for an effect from $X_i(t_1)$ to $X_j(t_2)$, which
is approximated linearly.
It is important to aggregate the absolute values of the coefficients to avoid cancellation
of positive and negative coefficients. The score $\widehat{\theta}_{ij}$ as such contains no information about whether the effect from $X_i(t_1)$ to $X_j(t_2)$
is positive or negative and it cannot be used to predict $X_j(t_2)$ from $X_i(t_1)$.
It serves as a score for the existence of a link between the two variables.
This rationale explains how linear methods may be employed for edge detection in non-linear settings without requiring extensions of Granger-type methods that explicitly model the non-linear dynamics and hence come with additional sample complexity~\citep{marinazzo2008kernel,marinazzo2011nonlinear,stramaglia2012expanding,stramaglia2014synergy}.

\section{Large regression coefficients may predict causal links better in practice than small p-values}\label{sec:coeffvspval}

This section aims at providing intuition behind two phenomena:
We observed a considerable drop in the accuracy of our edge predictions whenever
1) we normalised the data or
2)~used the T-statistics corresponding to testing the hypothesis of regression coefficients being zero to score edges instead of the coefficients' absolute magnitude.
While one could try to attribute these phenomena to some undesired artefact in the competition setup,
it is instructive to instead try to understand when exactly one would expect such behaviour.

We illustrate a possible explanation behind these phenomena and do so in an iid setting in favour of a clear exposition, while the intuition extends to settings of time series observations and our proposed algorithms.
The key remark is, that under comparable noise variances, the variables' marginal variances tend to increase along the causal ordering.
If data are observed at comparable scales---say sea level pressure in different locations measured in the same units---or at scales that are in some sense naturally relative to the true data generating mechanism, then  absolute regression coefficients may be preferable to T-test statistics.
Effect variables tend to have larger marginal variance than their causal ancestors.
This helpful signal in the data is diminished by normalising the data or the rescaling when computing the T-statistics corresponding to testing the regression coefficients for being zero.
This rationale is closely linked to the identifiability of Gaussian structural equation models under equal error variances~\citet{peters2014identifiability}.
Without any prior knowledge about what physical quantities the variables correspond to and their natural scales, normalisation remains a reasonable first step.
We are not advocating that one should use the raw coefficients and not normalise data, but these are two possible alterations of existing structure learning procedures that may or may not, depending on the concrete application at hand, be worthwhile exploring.
Our algorithms do not perform data normalisation, so the choice is up to the user whether to feed normalised or raw data, and one could easily change to using p-values or T-statistics instead of raw coefficients for edge scoring.

\subsection{Instructive iid case simulation illustrates scaling effects}\label{sec:iidsimulation}
We consider data simulated from a standard acyclic linear Gaussian model.
Let $N \sim \mathcal{N}\left(0, \operatorname{diag}(\sigma_1^2, \dots, \sigma_d^2)\right)$ be a $d$-dimensional random variable and let $\boldsymbol{B}$ be a $d\times d$ strictly lower-triangular matrix.
Further, let $X$ be a $d$-valued random variable constructed according to the structural equation $X = \boldsymbol{B} X + N$, which induces
a distribution over $X$ via $X = (I - \bm{B})^{-1}N$.
We have assumed, without loss of generality, that the causal order is aligned such that $X_i$ is further up in the causal order than $X_j$ whenever $i<j$.
We ran $100$ repetitions of the experiment, each time sampling a random lower triangular $50\times 50$-matrix $\bm{B}$ where each entry in the lower triangle is drawn from a standard Gaussian with probability $\sfrac{1}{4}$ and set to zero otherwise.
For each such obtained $\bm{B}$ we sample $n=200$ observations from $X=\bm{B}X+N$ which we arrange in a data matrix $\bm{X}\in\mathbb{R}^{200 \times 50}$ of zero-centred columns denoted by $\bm{X}_{j}$.

We regress each $X_j$ onto all remaining variables $X_{\neg j}$ and compare scoring edges $X_i\to X_j$ by the absolute values of a) the regression coefficients $|\widehat{b}_{i\to j}|$, versus b) the T-statistics $|\widehat{t}_{i\to j}|$ corresponding to testing the hypothesis that the regression coefficient $\widehat{b}_{i\to j}$ is zero. That is, we consider
\[
|\widehat{b}_{i\to j}| = \left|(\bm{X}_{\neg j}^\top \bm{X}_{\neg j})^{-1}\bm{X}_{\neg j}^\top \bm{X}_j\right|_i
\]
versus
\begin{equation}
|\widehat{t}_{i\to j}|  =
|\widehat{b}_{i\to j}|
{\color{factor}\sqrt{
\frac{
\widehat{\operatorname{var}}(X_i|X_{\neg i})
}{
\widehat{\operatorname{var}}(X_j|X_{\neg j})
}}}
\sqrt{\frac{
(n - d)
}{
\left(1 - \widehat{\operatorname{corr}}^2({X_i,X_j|X_{\neg \{i, j\}}})\right)
}}\label{eq:t_mulit}
\end{equation}
where
$
\widehat{\operatorname{var}}(X_j|X_{\neg j})$
is the residual variance after regressing $X_j$ onto the other variables $X_{\neg j}$,
and
$
\widehat{\operatorname{corr}}({X_i,X_j|X_{\neg \{i, j\}}})
$
is the residual correlation between $X_i$ and $X_j$ after regressing both onto the remaining variables.

\begin{figure}[h]
\resizebox{\textwidth}{!}{\input{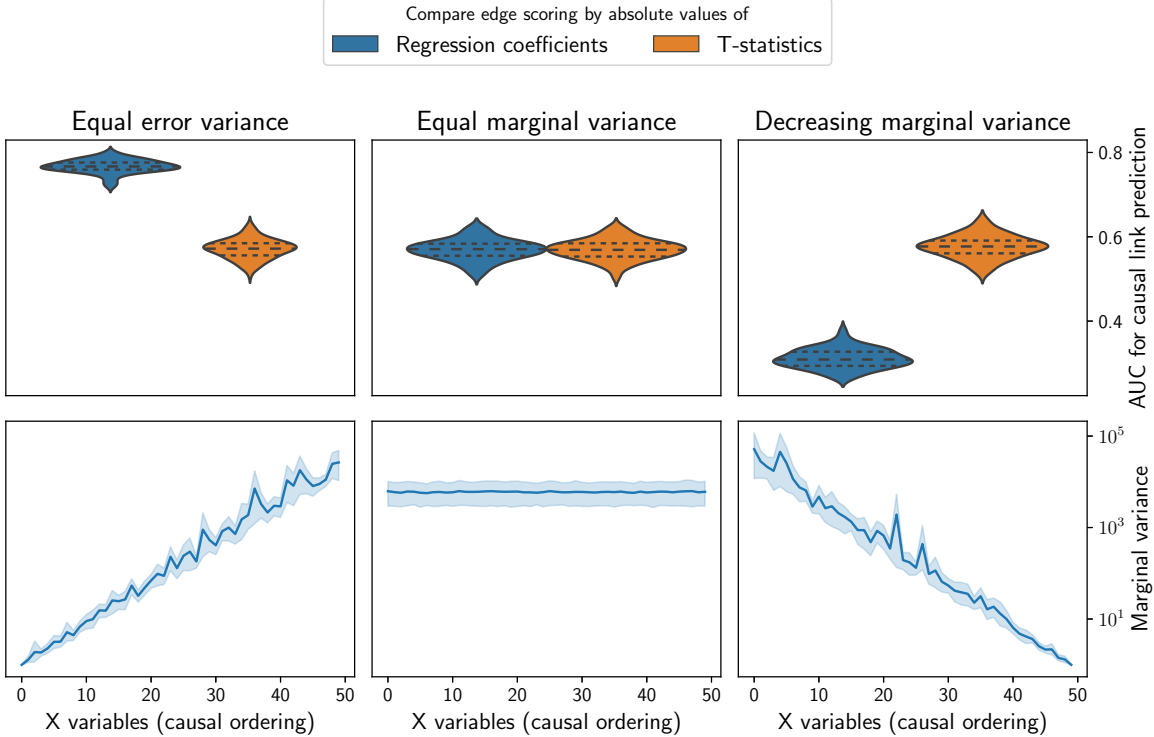}}
\caption{Results of the simulation experiment described in Section~\ref{sec:iidsimulation}.
Data is generated from an acyclic linear Gaussian model, in turn each variable is regressed onto all remaining variables and either the raw regression coefficient $|\widehat{b}_{i\to j}|$ or the corresponding T-statistics $|\widehat{t}_{i\to j}|$ is used to score the existence of an edge $i\to j$. %
The top row shows the obtained AUC for causal link prediction and the bottom row the marginal variance of the variables along the causal ordering.
The left panel shows naturally increasing marginal variance for equal error variances, for the middle and right panel the model parameters and error variances are rescaled to enforce equal and decreasing marginal variance, respectively.
}\label{fig:simulation}
\end{figure}

We now compare, across three settings, the AUC obtained by either using the absolute value of the regression coefficients $|\widehat{b}_{i\to j}|$ or the absolute value of the corresponding T-statistics $|\widehat{t}_{i\to j}|$ for edge scoring.
Results are shown in the left, middle, and right panel of Figure~\ref{fig:simulation}, respectively.

\paragraph{In the setting with equal error variances $\sigma_i^2 = \sigma^2_j\ \forall i,j$,}
we observe that
i) the absolute regression coefficients beat the T-statistics for edge predictions in terms of AUC, and
ii) the marginal variances naturally turn out to increase along the causal ordering.

When moving from $|\widehat{b}_{i\to j}|$ to $|\widehat{t}_{i\to j}|$ for scoring edges, we multiply by a term that compares the relative residual variance of $X_i$ and $X_j$.
If $X_i$ is before $X_j$ in the causal ordering it tends to have both smaller marginal and---in our simulation set-up---residual variance than $X_j$ as it becomes increasingly more difficult to predict variables further down the causal ordering.
In this case, the fraction of residual variances will tend to be smaller than one and consequently the raw regression coefficients $|\widehat{b}_{i\to j}|$ will be shrunk when moving to $|\widehat{t}_{i\to j}|$.
This can explain the worse performance of the T-statistics compared to the raw regression coefficients for edge scoring as scores will tend to be shrunk when in fact $X_i \to X_j$.

\paragraph{Enforcing equal marginal variances by rescaling the rows of $\bm{B}$ and the $\sigma_i^2\text{'s}$,}
we indeed observe that regression coefficients and T-statistics achieve comparable performance in edge prediction in this somewhat artificial scenario.
Here, neither the marginal variances nor the residual variances appear to contain information about the causal ordering any more and the relative ordering between regression coefficients and T-statistics is preserved when multiplying by the factor {\color{factor}highlighted} in Equation~\ref{eq:t_mulit}.

\paragraph{Enforcing decreasing marginal variances by rescaling the rows of $\bm{B}$ and the $\sigma_i^2\text{'s}$,}
we can, in line with our above reasoning, indeed obtain an artificial scenario in which the T-statistics will outperform the regression coefficients in edge prediction, as now, the factors we multiply by will work in favour of the T-statistics.

\section{Conclusion and Future Work}

We believe competitions like the Causality 4 Climate competition~\citep{runge2020neurips} and causal discovery benchmark platforms like \href{http://causeme.net}{CauseMe.net}~\citep{runge2019inferring} are important for bundling and informing the community's joint research efforts into methodology that is readily applicable to tackle real-world data.
In practice, there are fundamental limitations to causal structure learning that ultimately require us to employ untestable causal assumptions to proceed towards applications at all.
Yet, both these limitations and assumptions are increasingly well understood and characterised by methodological research and time and again need to be challenged and examined through the application to real-world data.

Beyond the algorithms presented here and proposed for baseline benchmarks, different methodology as well as different benchmarks may be of interest.
For example, our methods detect causal links and are viable benchmarks for the structure learning task but they do not per se enable predictions about the interventional distributions.

\acks{
The authors thank Niels Richard Hansen, Steffen Lauritzen, and Jonas Peters for insightful discussions.
Thanks to the organisers for a challenging and insightful Causality 4 Climate NeurIPS competition.
NT was supported by a research grant (18968) from VILLUM FONDEN.
LP and GV were supported by a research grant (13358) from VILLUM FONDEN.
MEJ and SW were supported by the Carlsberg Foundation.
}

\clearpage

\bibliography{references}

\begin{thebibliography}{23}
\providecommand{\natexlab}[1]{#1}
\providecommand{\url}[1]{\texttt{#1}}
\expandafter\ifx\csname urlstyle\endcsname\relax
  \providecommand{\doi}[1]{doi: #1}\else
  \providecommand{\doi}{doi: \begingroup \urlstyle{rm}\Url}\fi

\bibitem[Bongers and Mooij(2018)]{bongers2018random}
S.~Bongers and J.~M. Mooij.
\newblock {From random differential equations to structural causal models: The
  stochastic case}.
\newblock \emph{arXiv preprint arXiv:1803.08784}, 2018.

\bibitem[Danks and Plis(2013)]{danks2013learning}
D.~Danks and S.~Plis.
\newblock Learning causal structure from undersampled time series.
\newblock In \emph{JMLR: Workshop and Conference Proceedings}, 2013.

\bibitem[Granger(1969)]{granger1969investigating}
C.~W.~J. Granger.
\newblock Investigating causal relations by econometric models and
  cross-spectral methods.
\newblock \emph{Econometrica}, 37\penalty0 (3):\penalty0 424--438, 1969.

\bibitem[Hyttinen et~al.(2016)Hyttinen, Plis, J{\"a}rvisalo, Eberhardt, and
  Danks]{hyttinen2016causal}
A.~Hyttinen, S.~Plis, M.~J{\"a}rvisalo, F.~Eberhardt, and D.~Danks.
\newblock {Causal Discovery from Subsampled Time Series Data by Constraint
  Optimization}.
\newblock In \emph{Proceedings of the Eighth International Conference on
  Probabilistic Graphical Models}, 2016.

\bibitem[IPCC(2013)]{ipcc2013}
IPCC.
\newblock \emph{Climate Change 2013: The Physical Science Basis. Contribution
  of Working Group I to the Fifth Assessment Report of the Intergovernmental
  Panel on Climate Change}.
\newblock Cambridge University Press, 2013.

\bibitem[Janzing et~al.(2018)Janzing, Rubenstein, and
  Sch\"olkopf]{janzing2018structural}
D.~Janzing, P.~K. Rubenstein, and B.~Sch\"olkopf.
\newblock Structural causal models for macro-variables in time-series.
\newblock \emph{arXiv preprint arXiv:1804.03911}, 2018.

\bibitem[Marinazzo et~al.(2008)Marinazzo, Pellicoro, and
  Stramaglia]{marinazzo2008kernel}
D.~Marinazzo, M.~Pellicoro, and S.~Stramaglia.
\newblock {Kernel-Granger causality and the analysis of dynamical networks}.
\newblock \emph{Physical Review E}, 77\penalty0 (5):\penalty0 056215, 2008.

\bibitem[Marinazzo et~al.(2011)Marinazzo, Liao, Chen, and
  Stramaglia]{marinazzo2011nonlinear}
D.~Marinazzo, W.~Liao, H.~Chen, and S.~Stramaglia.
\newblock {Nonlinear connectivity by Granger causality}.
\newblock \emph{NeuroImage}, 58\penalty0 (2):\penalty0 330 -- 338, 2011.

\bibitem[Mogensen and Hansen(2020)]{wengel2019markov}
S.~W. Mogensen and N.~R. Hansen.
\newblock Markov equivalence of marginalized local independence graphs.
\newblock \emph{The Annals of Statistics}, 48\penalty0 (1):\penalty0 539--559,
  2020.

\bibitem[Mooij et~al.(2013)Mooij, Janzing, and Sch{\"o}lkopf]{mooij2013from}
J.~M. Mooij, D.~Janzing, and B.~Sch{\"o}lkopf.
\newblock {From Ordinary Differential Equations to Structural Causal Models:
  the deterministic case}.
\newblock In \emph{Proceedings of the Twenty-Ninth Conference on Uncertainty in
  Artificial Intelligence (UAI)}. AUAI Press, 2013.

\bibitem[Pearl(2009)]{Pearl2009}
J.~Pearl.
\newblock \emph{Causality}.
\newblock Cambridge University Press, 2 edition, 2009.

\bibitem[Peters and B{\"u}hlmann(2014)]{peters2014identifiability}
J.~Peters and P.~B{\"u}hlmann.
\newblock Identifiability of {G}aussian structural equation models with equal
  error variances.
\newblock \emph{Biometrika}, 101\penalty0 (1):\penalty0 219--228, 2014.

\bibitem[Peters et~al.(2017)Peters, Janzing, and
  Sch\"olkopf]{peters2017elements}
J.~Peters, D.~Janzing, and B.~Sch\"olkopf.
\newblock \emph{Elements of Causal Inference}.
\newblock MIT Press, 2017.

\bibitem[Peters et~al.(2020)Peters, Bauer, and Pfister]{Peters2020}
J.~Peters, S.~Bauer, and N.~Pfister.
\newblock Causal models for dynamical systems.
\newblock \emph{arXiv preprint arXiv:2001.06208}, 2020.

\bibitem[Rubenstein et~al.(2017)Rubenstein, Weichwald, Bongers, Mooij, Janzing,
  Grosse-Wentrup, and Sch{\"o}lkopf]{rubenstein2017causal}
P.~K. Rubenstein, S.~Weichwald, S.~Bongers, J.~M. Mooij, D.~Janzing,
  M.~Grosse-Wentrup, and B.~Sch{\"o}lkopf.
\newblock Causal consistency of structural equation models.
\newblock In \emph{Proceedings of the Thirty-Third Conference on Uncertainty in
  Artificial Intelligence (UAI)}. AUAI Press, 2017.

\bibitem[Rubenstein et~al.(2018)Rubenstein, Bongers, Mooij, and
  Sch{\"o}lkopf]{rubenstein2016from}
P.~K. Rubenstein, S.~Bongers, J.~M. Mooij, and B.~Sch{\"o}lkopf.
\newblock From deterministic {ODE}s to dynamic structural causal models.
\newblock In \emph{Proceedings of the 34th Annual Conference on {U}ncertainty
  in {A}rtificial {I}ntelligence ({UAI})}. AUAI Press, 2018.

\bibitem[Runge et~al.(2019{\natexlab{a}})Runge, Bathiany, Bollt, Camps-Valls,
  Coumou, Deyle, Glymour, Kretschmer, Mahecha, Mu{\~n}oz-Mar{\'\i}, Peters,
  Quax, Reichstein, Scheffer, Sch\"olkopf, Spirtes, Sugihara, Sun, Zhang, and
  Zscheischler]{runge2019inferring}
J.~Runge, S.~Bathiany, E.~Bollt, G.~Camps-Valls, D.~Coumou, E.~Deyle,
  C.~Glymour, M.~Kretschmer, M.~D. Mahecha, E.~H. Mu{\~n}oz-Mar{\'\i}, J.
  andand van~Nes, J.~Peters, R.~Quax, M.~Reichstein, M.~Scheffer,
  B.~Sch\"olkopf, P.~Spirtes, G.~Sugihara, J.~Sun, K.~Zhang, and
  J.~Zscheischler.
\newblock {Inferring causation from time series in Earth system sciences}.
\newblock \emph{Nature Communications}, 10\penalty0 (1):\penalty0 2553,
  2019{\natexlab{a}}.

\bibitem[Runge et~al.(2019{\natexlab{b}})Runge, Nowack, Kretschmer, Flaxman,
  and Sejdinovic]{runge2019detecting}
J.~Runge, P.~Nowack, M.~Kretschmer, S.~Flaxman, and D.~Sejdinovic.
\newblock Detecting and quantifying causal associations in large nonlinear time
  series datasets.
\newblock \emph{Science Advances}, 5\penalty0 (11), 2019{\natexlab{b}}.

\bibitem[Runge et~al.(2020)Runge, Tibau, Bruhns, Mu{\~n}oz-Mar\'i, and
  Camps-Valls]{runge2020neurips}
J.~Runge, X.-A.~Tibau, M.~Bruhns, J.~Mu{\~n}oz-Mar\'i, and
  G.~Camps-Valls.
\newblock The causality for climate competition.
\newblock In Hugo~Jair Escalante and Raia Hadsell, editors, \emph{{PMLR NeurIPS
  Competition \& Demonstration Track Postproceedings}}, Proceedings of Machine
  Learning Research. PMLR, 2020.
\newblock URL \url{https://causeme.uv.es/}.
\newblock Forthcoming.

\bibitem[Spirtes et~al.(2001)Spirtes, Glymour, and Scheines]{Spirtes2000}
P.~Spirtes, C.~N. Glymour, and R.~Scheines.
\newblock \emph{{Causation, Prediction, and Search}}.
\newblock MIT Press, 2 edition, 2001.

\bibitem[Stramaglia et~al.(2012)Stramaglia, Wu, Pellicoro, and
  Marinazzo]{stramaglia2012expanding}
S.~Stramaglia, G.-R. Wu, M.~Pellicoro, and D.~Marinazzo.
\newblock Expanding the transfer entropy to identify information circuits in
  complex systems.
\newblock \emph{Physical Review E}, 86\penalty0 (6):\penalty0 066211, 2012.

\bibitem[Stramaglia et~al.(2014)Stramaglia, Cortes, and
  Marinazzo]{stramaglia2014synergy}
S.~Stramaglia, J.~M. Cortes, and D.~Marinazzo.
\newblock {Synergy and redundancy in the Granger causal analysis of dynamical
  networks}.
\newblock \emph{New Journal of Physics}, 16\penalty0 (10):\penalty0 105003,
  2014.

\bibitem[Wiener(1956)]{wiener1956theory}
N.~Wiener.
\newblock The theory of prediction.
\newblock \emph{Modern Mathematics for Engineers}, 1956.

\end{thebibliography}

\end{document}